\documentclass[a4paper]{article}
\usepackage{CJKutf8}
\usepackage{amsmath}
\usepackage{INTERSPEECH2020}
\usepackage{xcolor}
\usepackage{tipa}

\title{Improving Code-switching Language Modeling with Artificially Generated Texts using Cycle-consistent Adversarial Networks}
\name{Chia-Yu Li, Ngoc Thang Vu}
\address{Institute for Natural Language Processing (IMS), University of Stuttgart, Germany}
\email{\{licu, thangvu\}@ims.uni-stuttgart.de}

\begin{document}

\maketitle
\begin{abstract}
  This paper presents our latest effort on improving Code-switching language models that suffer from data scarcity. We investigate methods to augment Code-switching training text data by artificially generating them. Concretely, we propose a cycle-consistent adversarial networks based framework to transfer monolingual text into Code-switching text, considering Code-switching as a speaking style. Our experimental results on the SEAME corpus show that utilizing artificially generated Code-switching text data improves consistently the language model as well as the automatic speech recognition performance.
\end{abstract}

\section{Introduction}

Code-switching (CS) speech is a common phenomenon in multilingual countries and defined as speech that contains more than one language \cite{auer2013code}. Recently, CS speech has received great attention in speech communities and is identified as one of the most challenging tasks for automatic speech recognition (ASR) systems, either for hybrid systems (e.g.\cite{kaldi}) or end-to-end systems (e.g.\cite{espnet}). One of the main challenges is the data scarcity issue, especially in the context of language modeling. 

Data augmentation provides a potential solution for data scarcity because it is less time-consuming than collecting and transcribing real speech data and people have shown in many contexts \cite{ko2015audio, gorin2016language, shorten2019survey, park2019specaugment, Bao2019} that it improves results. Text generation - a data augmentation method - has been proposed in \cite{1stCS, Vu2014Generation, Yilmaz2018, Winata2018, Chang2019, Gao2019, Lee2019} with the aim of improving language models and therefore automatic speech recognition performance for CS speech. The first approach \cite{1stCS, Chang2019, Gao2019, Lee2019} breaks down the main task in two consecutive steps: first to predict CS start and end points and then to replace a certain number of words or phrases in monolingual text with translated words or phrases in another language using either rules or machine learning approaches. This approach suffers from the error propagation problem. % - the main weakness of a pipeline architecture. 
The second approach \cite{Vu2014Generation, Yilmaz2018} leverages recurrent neural network language models trained with CS transcriptions. Because of the small amount of CS transcriptions, the generated texts are often meaningless \cite{Vu2014Generation}. The third approach \cite{Winata2018} formulates the CS text generation task as a sequence to sequence problem, mapping monolingual text to CS text and solves the task in an end-to-end fashion using sequence-to-sequence model \cite{S2S,NMT-S2S}. This model, however, requires a large amount of parallel training data. Moreover in other contexts of text generation, it tends to generate ill-formed text \cite{li-etal-2016-diversity, luo-etal-2018-auto}.

%Another option is to use generative adversarial networks (GANs) \cite{goodfellow2014generative} for text generation \cite{zhang2017adversarial, yu2017seqgan, xu-etal-2018-diversity}. 
Recently, CycleGAN - a variation of generative adversarial networks (GANs) \cite{goodfellow2014generative} for style transfer - has been proposed for translating images~\cite{CycleGAN2017} and then applied in other domains ~\cite{kaneko2018parallel,Bao2019}. The main idea of this model is to transfer one style of the source domain to another style of the target domain by learning a mapping between them without any parallel training data. The first attempt on text was done in ~\cite{fu2018style} in which the authors interpreted text styles in a common sense and proposed model to modify sentiment of the text and to transform the text title between paper and news styles. 

In this work, we propose the novel idea of considering Code-switching as a speaking style and therefore to use CycleGANs~\cite{CycleGAN2017} to generate artificially text data for two reasons: first, CycleGAN transfers styles without any parallel training data; second by doing so, CS text generation can be solved in an end-to-end fashion, meaning that the two tasks - CS points prediction and short text translation - will be jointly optimized to mimic the CS distributions in the real CS data and to maintain the same meaning when translating from one language to another. In order to successfully generate artificially CS text, we explore tricks how to train this complex model by using sequence-to-sequence model and by leveraging monolingual data, and investigate the impact of the cycle losses in CycleGAN on the language model and the automatic speech recognition performance.
%That means the same content could be linguistically realized in one language, e.g. either English or Chinese, or in a CS manner by mixing several languages. 
%Furthermore, CS text can be easily generated for any domain, thus building a large synthetic CS text corpus for CS language modeling \cite{adel2013recurrent}.

In sum, our contributions are as follows: 1) To the best of our knowledge, we are the first to propose a novel framework based on the CycleGAN architecture for generating artificially CS texts from monolingual texts for language modeling; 2) We show that our CS generated texts contribute to improve not only the language modeling performance but also the ASR system on the SEAME corpus.

\section{Proposed Method}
%In this section, we describe our novel methods to generate CS text from monolingual text using the sequence-to-sequence architecture and the CycleGAN technique.
%\vspace{-5mm}
\subsection{Sequence-to-sequence (S2S) Model}
The sequence-to-sequence model \cite{S2S} proposed an end-to-end approach to sequence learning that makes minimal assumptions on the sequence structure. 
%Sutskever's method uses an RNN encoder with Long Short-Term Memory units (LSTM) to map an input sequence to a vector of a fixed dimensionality, and then uses a decoder with the same structure to generate an output sequence from this vector. It has shown impressive results in the field of machine translation \cite{S2S,NMT-S2S}, in which the model takes a sequence of symbols of a source language as input and predicts a sequence of symbols of a target language. 
%If CS is considered as a language, then generating CS text from Chinese text could be seen as a machine translation problem. 
In the CS context, the S2S model takes a Chinese sentence as input and generates the Chinese-English CS sentence. For example, given a sequence of inputs (Chinese sentence) $(x_1,x_2,...x_T)$, the goal of the S2S models is to estimate the conditional probability:

\begin{equation}
%\footnotesize
    P(y_1,...,y_{T'}|x_1,...,x_T)=\prod_{i=1}^{T'} P(y_t|v,y_1,...,y_{t-1})
\end{equation}

where $(y_1,...,y_{T'})$ is the corresponding output sequence (Chinese-English CS sentence) whose length $T'$ may differ from $T$. $v$ is the representation of the input sequence $(x_1, . . . , x_T)$ given by the last hidden state of the encoder. The model is trained in a supervised manner. In CS, the trick is to generate paired training data by utilizing Chinese texts and their partial dictionary-based translations that form artificially generated CS texts \cite{1stCS}. The idea of this simple method is to replace a certain number of words or phrases in monolingual texts with their translations based on a dictionary, or, as in this work, using the Google Translation API. Our intuition is that the S2S model will learn more generalized patterns in CS behaviours by taking advantage of the continuous latent semantic space and thereby generate more CS variations than the dictionary based method.

\subsection{CycleGAN-based Model}
\begin{figure}[t]
    \centering
    \includegraphics[width=\linewidth]{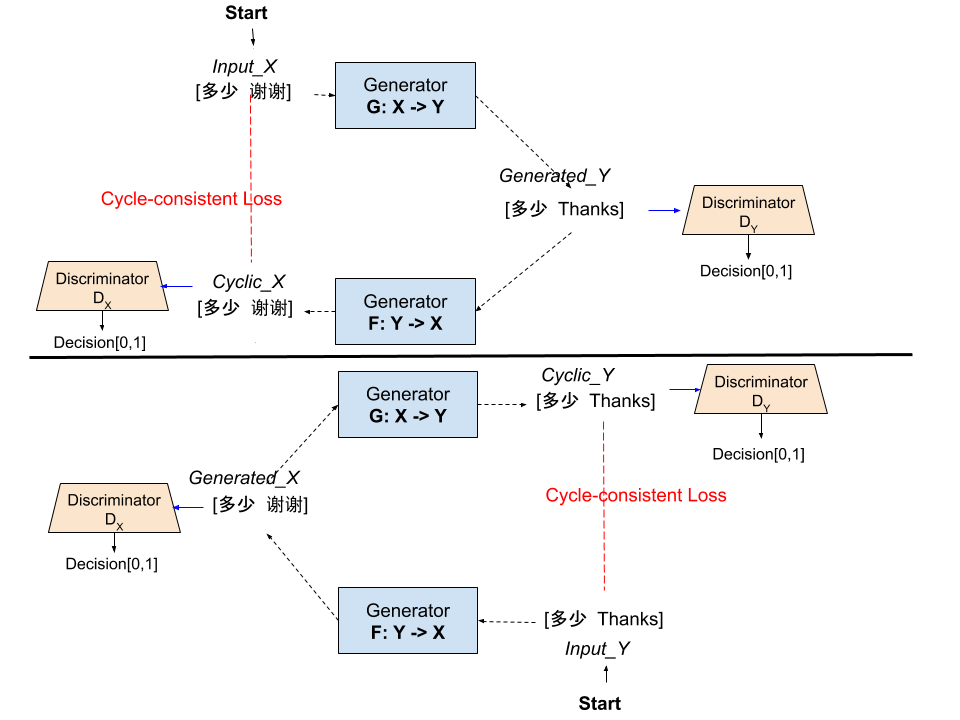}
    \caption{Network architecture of CycleGAN for CS text generation. X refers to monolingual text, e.g. Chinese and Y refers to CS text, e.g. Chinese English CS sentences.}
    \label{fig:cyclegan}
\end{figure}

The main disadvantage of the previous proposed method is that it does not make use of any prior knowledge encoded in an existing CS corpus. Therefore, we propose to employ the pretrained S2S models, which are $G: X \rightarrow Y$ and $F: Y \rightarrow X$ in Figure \ref{fig:cyclegan}, in the CycleGAN architecture to take advantage of both methods: the S2S generation framework and the CS prior knowledge encoded in a 'Discriminator' without any real paired data (monolingual text and its corresponding CS text). 

CycleGAN is a currently popular technique to solve image to image translation problems where the goal is to learn a mapping $G$ between an input image from a source domain $X$ and an output image from a target domain $Y$ without using paired training data \cite{CycleGAN2017}. 
%Zhu et al presented this approach to translate an image from a source domain $X$ to a target domain $Y$ in the absence of paired examples in order to learn a 
The mapping $G: X \rightarrow Y$ is learnt such that the distribution of images from $G(X)$ is indistinguishable from the distribution $Y$ using an adversarial loss. 
%This mapping is learnt such that the distribution of images from $G(X)$ is indistinguishable from the distribution $Y$ using an adversarial loss.
Because this mapping is highly under-constrained, \cite{CycleGAN2017} coupled it with an inverse mapping $F: Y \rightarrow X$ and introduce a cycle consistency loss to push $F(G(X))=X$ (and vice versa). The full objective is 

\begin{equation}
%\footnotesize
\begin{aligned}
    \mathcal{L}(G,F,D_X,D_Y)=\mathcal{L}_{GAN}(G,D_Y,X,Y)\\+\mathcal{L}_{GAN}(F,D_X,Y,X)+\lambda\mathcal{L}_{cyc}(G,F)
\end{aligned}
\end{equation}

where $G$ tries to generate images $G(x)$ that look similar to images from domain $Y$ , while $D_Y$ aims to distinguish between translated samples $G(x)$ and real samples $y$. The same holds for the Discriminator $D_X$ which tries to distinguish between $F(Y)$ and real $x$. $\lambda$ controls the relative importance of cycle consistency loss:
\begin{equation}
%\footnotesize
\begin{aligned}
    \mathcal{L}_{cyc}(G,F)=\mathbf{E}_{x\sim p_{data}(x)}[\mid\mid F(G(x))-x \mid\mid_1]\\
    +\mathbf{E}_{y\sim p_{data}(y)}[\mid\mid G(F(y))-y \mid\mid_1]
\end{aligned}
\end{equation}
In the context of CS, X refers to monolingual text (e.g. Chinese text) and Y refers to CS text. We define that 
\begin{equation}
%\footnotesize
\begin{aligned}
\mathcal{L}_{cyc}(F(G(X))) =\mathbf{E}_{x\sim p_{data}(x)}[\mid\mid F(G(x))-x \mid\mid_1]\\
\mathcal{L}_{cyc}(G(F(Y))) = \mathbf{E}_{y\sim p_{data}(y)}[\mid\mid G(F(y))-y \mid\mid_1] 
\end{aligned}
\end{equation}
Furthermore, we adapt the objective to the following equation:
\begin{equation}
%\footnotesize
\begin{aligned}
    \mathcal{L}(G,F,D_X,D_Y)=\mathcal{L}_{GAN}(G,D_Y,X,Y)
    \\+\mathcal{L}_{GAN}(F,D_X,Y,X)
    \\+\lambda_1\mathcal{L}_{cyc}(F(G(X)))
    \\+\lambda_2\mathcal{L}_{cyc}(G(F(Y)))
\end{aligned}
\label{eq:loss}
\end{equation}
where $\lambda\_1$ and $\lambda\_2$ are tunable parameters to control the importance of the two cycle consistency losses. Moreover, we can tune the balance between them.

Figure \ref{fig:cyclegan} shows the network architecture of CycleGAN for generation of CS text. There are two Generators $G$ and $F$: one transforms a Chinese sentence to a CS one, another converts a CS sentence to a Chinese one. The goal of the two Discriminators ($D_X$ and $D_Y$) is to predict whether the sample is from the actual distribution ( `real') or produced by the Generator (`fake'). %For example, if the input for ZH discriminator (i.e. $D_X$) is generated\_X from $G$ generator, then it should predict 0. Otherwise, if the input is the sentence from real data (e.g. AISHELL-1), then it should predict 1. 
The constraint of cycle consistency enforces the network to use shared latent features that can be used to map this output back to the original input.
It could prevent the Generator from generating the same sentence in the target domain while the input is quite different. The upper part of Figure \ref{fig:cyclegan} describes how we compute the cycle consistency loss $\mathcal{L}_{cyc}(F(G(X)))$ and the lower part is for $\mathcal{L}_{cyc}(G(F(Y)))$.

\section{Resources}
\label{sec:seame}
\subsection{SEAME Dataset}
We use SEAME - the Mandarin-English Code-switching spontaneous speech corpus in South East Asia that contains 99 hours of spontaneous Chinese-English speech recorded from Singaporean and Malaysian. All recordings are done with close-talk microphones in a quiet room. The speakers are aged between 19 and 33 and almost balanced in gender (49.7\% female and 50.3 \% male). 16.96\% of utterances are English, 15.54\% are Chinese and the rest (67\%) are CS utterances \cite{seame}.%The total number of distinct speakers is 157 (36.8\% are Malaysian while the rest are Singaporean) \cite{seame}. 16.96\% of utterances are English, 15.54\% are Chinese and the rest (67\%) are CS utterances.
\subsection{Code-mixing Index (CMI)}
%Code-mixing index
In order to compare the artificially generated CS sentences with the real CS corpus, we use Code-mixing Index (CMI) at the utterance level introduced in \cite{CMI} as a measurement of the level of mixing between languages, that is
\begin{equation}
%\footnotesize
%\[
    CMI= 
\begin{cases}
    100 \times [1 - \frac{\max\{w_i\}}{(n-u)}], & n>u\\
    0,              & n=u
\end{cases}
%\]
\end{equation}
where $n$ is the total number of tokens, $u$ is the number of other non-verbal tokens (e.g. people laughing) and $\max\{w_i\}$ is the number of tokens from the dominant language.
If an utterance only contains non-verbal tokens (i.e., if $n=u$), it has an index of zero. %Otherwise, it normalizes the complement fraction of the most prominent language in the utterance. 
The CMI ranges from $[0,50]$. The lower the value, the lower the level of mixing between languages. Because this index does not indicate the dominant language in the utterances, we add the dominant language tag before the CMI and group them into ten groups presented in Table \ref{tab:cmi_propotion}. It illustrates ten groups of different mixing levels and the proportion of each group in the SEAME data set. 
%For example, ZH-C1 means that the dominant language in an utterance is Mandarin Chinese (ZH) and the CMI equals 0. 
For example, EN-C2 presents a portion of data with English as the dominant language and the CMI ranges (0,15]. It is 18\%, 16\% and 15\% of the SEAME train, dev and eval set, respectively.
%In the SEAME train set (without monolingual sentences), there are 18\%, 25\%, 21\% and 8\% of utterances belonging to group ZH-C2, ZH-C3, ZH-C4 and ZH-C5, respectively.  

\begin{table}[t]
    %\footnotesize
    \caption{Different levels of code-mixing in the SEAME corpus.}
    \label{tab:cmi_propotion}
    \centering
    \begin{tabular}{lcccc}
    \toprule
    \textbf{Groups}&\textbf{CMI}&\textbf{train (\%)}&\textbf{dev (\%)}&\textbf{eval (\%)} \\
    \midrule
        ZH-C1 &0&  1&1&3\\
        ZH-C2 &(0,15]&  18&16&15\\
        ZH-C3 &(15,30]&  25&26&24\\
        ZH-C4 &(30,45]&  21&23&20\\
        ZH-C5 &(45,50]&  8&9&9\\
        \hline
        EN-C1 &0&  0&0&0\\
        EN-C2 &(0,15]&  4&3&3\\
        EN-C3 &(15,30]&  9&9&11\\
        EN-C4 &(30,45]&  13&11&14\\
        EN-C5 &(45,50]&  1&1&1\\
    \bottomrule
    \end{tabular}
\end{table}

\section{Experimental setup}
\label{sec:exp}
\subsection{Monolingual Datasets}
We utilize two Chinese datasets to generate Code-switching text which were used to train the S2S RNN model and the pretrained generators in the CycleGAN:
1) Aishell-1 contains 96,078 sentences contributed by 400 people in China and covers “Finance”, “Science and Technology”, “Sports”, “Entertainments” and “News” topics \cite{aishell-1}.
2) The National University of Singapore SMS Corpus is a corpus of Short Message Service (SMS) messages collected for research at the Department of Computer Science at the National University of Singapore. It consists of 29,031 SMS messages in Chinese and the messages largely originate from Singaporeans and mostly from students attending the University \cite{SMS}.

\subsection{Baseline Systems}
%We evaluate our proposed method in the language modeling tasks and their impact on speech recognition performance. 
The baseline language model is trained with SEAME training text containing 94,504 sentences that includes monolingual (English or Chinese) and CS sentences. The word-based language model is one layer of LSTM with 650 units and trained with a batch size of 20. It achieves perplexities of 84 and 75 on “dev” and “eval” set, respectively. The subword-based language model is a two layers RNN model with 650 units and trained with the batch size of 256 (Espnet default settings). It has a perplexity of 12.37 on “dev” and 10.90 on “eval” set. 
The baseline end-to-end (E2E) ASR is trained by Espnet \cite{espnet} using our best recipe presented in \cite{espnetcs}. The mixed error rate (MER), which is the error rate on Chinese characters and English words, on SEAME “dev” and “eval” set are 30.7\% and 23.0\% without fusing any external language model, respectively. Integrating the baseline language model to our E2E ASR using shallow fusion results in a MER of 30.4\% on “dev” and 22.4\% on “eval” set.
%Table \ref{tab:wer} and \ref{tab:wer2}.

\subsection{Implementation \& Hyperparameters}
The S2S and CycleGAN models are implemented using Python3 and Pytorch. %\cite{pytorch}. 
Both S2S and the two Generators in CycleGAN use RNN encoder decoder that consists of two RNNs with 650 LSTM units. 
%One RNN encodes a sequence of symbols into a fixed-length vector representation, and the other decodes the representation into another sequence of symbols. This model can alleviate the problem of information compression while using multi-layers LSTM Encoder-Decoder \cite{NMT-S2S}. 
The models are trained with 96,078 Chinese sentences from Aishell-1 and their corresponding CS artificially generated sentences.
The two Discriminators in CycleGAN are trained with SEAME training data containing 94,504 sentences and 96,078 Chinese sentences from the Aishell-1 dataset.
The dimension of word embeddings is 300 and the batch size is 64. The optimizer is Adam \cite{adam}. The best tunable parameters ($\lambda\_1$ and $\lambda\_2$) in the objective of CycleGAN on the development set are 0.3 and 0.8, respectively.

\section{Results}
\label{sec:result}
\subsection{Language Model Evaluation}
Table \ref{tab:perplexity} shows the perplexity of two different language models (word-based LM and subword-based LM) on SEAME “dev” and “eval” set. We use the subword-based LM because our best E2E ASR system \cite{espnetcs} outputs subwords. Both are trained with three different texts: SEAME training text, SEAME text adding the text from S2S model, and SEAME text adding the text from CycleGAN. The results show that the language model trained with SEAME and the generated text from CycleGAN outperforms the baseline LM and the one trained with SEAME and the generated text from S2S model on the “eval” set. 

\begin{table}[!ht]
    %\footnotesize
    \caption{The PPL on SEAME “dev” and “eval” set}
    \label{tab:perplexity}
    \centering
    \begin{tabular}{l|c|c}
    \hline
    %& \multicolumn{2}{c}{\textbf{w/o mono}} &\multicolumn{2}{|c}{\textbf{with mono}}\\
    \textbf{Word-based LM} & \textbf{dev} & \textbf{eval}\\
    \hline
    SEAME & 84.00 & 75.00   \\
    %\hline
    +text from S2S & \textbf{83.03} &  74.67  \\
    %\hline
    +text from CycleGAN &  83.06 & \textbf{72.59} \\
    \hline
    \hline
    \textbf{Subword-based LM} & \textbf{dev} & \textbf{eval}\\
    \hline
    SEAME & 12.37 & 10.90\\
    %\hline
    +text from S2S & 12.13 &10.80  \\
    %\hline
    +text from CycleGAN &  \textbf{10.39} &  \textbf{9.59}\\
    \hline
    \end{tabular}
\end{table}

\subsection{End-to-End Speech Recognition Evaluation}
The subword-based LM is integrated to the E2E speech recognition system using shallow fusion \cite{espnetcs}, and the MERs on SEAME “dev” and “eval” set are presented in Table \ref{tab:wer} and \ref{tab:wer2}. `Overall' means the entire “dev” or “eval” set, and `EN', `ZH' and `CS' are denoted as the monolingual English sentences, monolingual Chinese sentences and Chinese-English CS sentences in “dev” or “eval” set, respectively. The results on both sets show that the MER is improved most not only on CS utterances but also on monolingual ones when the language model trained with the combination of SEAME text and the generated CS text from CycleGAN.
%On the dev set, the LM improves the performance on monolingual and CS test sets. 

\begin{table}[!htb]
    %\footnotesize
    \caption{The MER on SEAME “dev” set}
    \label{tab:wer}
    \centering
    \begin{tabular}{l|c|c|c|c}
    \hline
    \textbf{Subword-based LM} & \textbf{Overall} & \textbf{EN} & \textbf{ZH} & \textbf{CS} \\
    \hline
    SEAME & 30.4 & 43.4 & 23.8 & 30.3 \\
    %\hline
    +text from S2S & 30.3 & 42.8  & 23.4 & 30.2 \\
    %\hline
    +text from CycleGAN & \textbf{30.1}  & \textbf{42.6} & \textbf{23.2} & \textbf{30.0} \\
    \hline
    \end{tabular}
\end{table}

\begin{table}[!htb]
   %\footnotesize
   \caption{The MER on SEAME “eval” set}
    \label{tab:wer2}
    \centering
    \begin{tabular}{l|c|c|c|c}
    \hline
    \textbf{Subword-based LM} & \textbf{Overall} & \textbf{EN} & \textbf{ZH} & \textbf{CS} \\
    \hline
    SEAME & 22.4 & 32.4 & 22.7& 21.5 \\
    %\hline
    +text from S2S & 22.1 & 33.0 & 23.3 & 21.2\\
    %\hline
    +text from CycleGAN & \textbf{21.9} & \textbf{32.5} & \textbf{22.1} & \textbf{21.0} \\
    \hline
    \end{tabular}
\end{table}

\section{Quantitative \& Qualitative Analysis}
\label{sec:analysis}
\subsection{Impact of $\lambda\_1$ and $\lambda\_2$} 
Table \ref{tab:diff_lambda_dev} shows the perplexity of subword-based language model on SEAME dev sets with different values of $\lambda\_1$ and $\lambda\_2$.% The models with $\lambda_1=0.1$ and $\lambda_1=0.2$  are worse than the ones with other values of $\lambda_1$. 
$\lambda\_1=0.3$ seems to deliver the best performance in most cases.
The E2E systems that integrate language models trained with text generated usng CycleGAN framework either with $\lambda\_1=0.3$ and $\lambda\_2=0.8$ or with $\lambda\_1=0$ and $\lambda\_2=0.6$ achieve the best MER on both SEAME “dev” and “eval” set.

\begin{table}[!htb]
    %\footnotesize
    \centering
    \caption{The PPL of subword-based LM on SEAME “dev” set}
    \begin{tabular}{c|c|c|c|c|c}
    \hline
    \boldsymbol{$\lambda_2$}  $\backslash$  \boldsymbol{$\lambda_1$}   &  \bf{0} & \bf{0.1} & \bf{0.2} & \bf{0.3} & \bf{0.4} \\
    \hline
       0.5 &  11.56 & 12.09 & 12.18 & 11.11 & \bf{10.62}\\
       0.6 &  11.02  & 11.14 & 11.66 & \bf{10.81} & 11.64\\
       0.7 &  11.05 & 12.22 & 10.57 & \bf{10.56} & 11.07\\
       0.8 & 11.94 & 11.82&11.87&\bf{10.39}&11.62\\
       0.9 & \bf{10.43} & 11.23&11.12&11.96&10.77\\
       \hline
    \end{tabular}
    \label{tab:diff_lambda_dev}
\end{table}

\begin{table}[!htb]
    %\footnotesize
    \caption{Different levels of code-mixing in the SEAME text and, in the texts generated from S2S and CycleGAN.}
    \label{tab:cmi_propotion_synthetic}
    \centering
    \begin{tabular}{l|c|c|c|c}
    \hline
    \textbf{Groups}&\textbf{CMI}&\textbf{SEAME}&\textbf{S2S}&\textbf{CycleGAN} \\
    &&\textbf{(\%)}&\textbf{(\%)}&\textbf{(\%)}\\
    \hline
        ZH-C1 &0    &  1&0&0\\
        ZH-C2 &(0,15]&  17&13&12\\
        ZH-C3 &(15,30]&  25&12&20\\
        ZH-C4 &(30,45]&  21&12&22\\
        ZH-C5 &(45,50]&  9&13&5\\
        \hline
        EN-C1 &0    &  0&0&2\\
        EN-C2 &(0,15]&  4&13&10\\
        EN-C3 &(15,30]&  9&12&11\\
        EN-C4 &(30,45]&  13&13&18\\
        EN-C5 &(45,50]&  1&12&0\\
    \hline
    \end{tabular}
\end{table}

\subsection{Generated Text} 
We analyzed 148,853 generated CS sentences from S2S and CycleGAN and show some interesting examples in this subsection. Table \ref{tab:cmi_propotion_synthetic} shows 
that when comparing the distribution of CMI groups, texts generated by CycleGAN have a more similar distribution to the ones in SEAME than texts generated by S2S across all different levels of code-mixing. This fact suggests that CycleGAN successfully enforces the generated texts to mimic CS behaviour in the original corpus.

\begin{CJK*}{UTF8}{gbsn}
\begin{table}[ht]
    \footnotesize
    \caption{The output from S2S and CycleGAN models}
    \label{tab:example}
    \centering
    \begin{tabular}{l|l}
    \hline
    \bf{Monolingual}&提高\hspace{5mm}铁路\hspace{3mm}在\hspace{1mm}物流\hspace{5mm}市场\hspace{4mm}中\hspace{1mm}的\hspace{1mm}竞争力\\
      \hline
      S2S&improve\hspace{1mm}铁路\hspace{3mm}在\hspace{1mm}logistics\hspace{1mm}market\hspace{2mm}in\hspace{1mm}的\hspace{1mm}竞争力\\
      CycleGAN&improve\hspace{1mm}railway\hspace{1mm}in\hspace{1mm}\textbf{2016}\hspace{5mm}market\hspace{2mm}in\hspace{1mm}的\hspace{1mm}竞争力\\
      \hline
    \end{tabular}
\end{table}
\end{CJK*}

\begin{CJK*}{UTF8}{gbsn}
\begin{table}[h]
    %\footnotesize
    \caption{CycleGAN simulates stuttering speaking style}
    \label{tab:example3}
    \centering
    \begin{tabular}{l}
    \toprule
    \textbf{Reference (NI01MAX0101)}\\
    \hline
      \textbf{take take}\hspace{1mm}多 \hspace{1mm}一个\hspace{1mm}then\hspace{1mm}就 then\hspace{1mm}就\hspace{1mm}这\\
      but\hspace{1mm}那\hspace{1mm}边\hspace{1mm}的\hspace{1mm}东西\hspace{1mm}最好\hspace{1mm}了我\hspace{1mm}觉得\hspace{1mm}in terms \textbf{of of}\hspace{1mm}那\\
      \hline
      \hline
      \textbf{CycleGAN}\\
      \hline
      and \hspace{1mm}如果\hspace{1mm} \textbf{take take} out of ten \\
      根据\hspace{1mm}one series \textbf{of of of}\hspace{1mm}问题\\
    \bottomrule
    \end{tabular}
\end{table}
\end{CJK*}

While we aim at generating texts with the same meaning of the inputs, we observed an interesting effect as shown in an example in Table \ref{tab:example}. While the output of S2S contains Chinese words being replaced by their translations, CycleGAN produces words with different meanings ("2016" marked in bold instead of the English noun "logistics"). 
In this case, CycleGAN transforms the input sentence into a different meaningful CS one instead of just replacing Chinese words by their translations.

Furthermore, we observe that text generated using CycleGAN framework might simulate speaking styles used in SEAME. Table \ref{tab:example3} presents two examples in which CycleGAN generates text with some repetitive words, mimicking real speakers in SEAME data who tends to stutter. 

\subsection{Analysis of Recognition Results} 
The top of Table \ref{tab:analysis} presents the recognition performance comparison in terms of average substitution, deletion, insertion and MER between the baseline and the improved ASR systems. 
The baseline system has less deletion, while adding artificially generated CS text from CycleGAN results in better substitution and insertion rates and the overall MER. 
We show some cherry picked ASR output examples in the bottom of Table \ref{tab:analysis} that show the positive impact of using artificially generated CS text data on improving the ASR performance in both monolingual (English and Chinese) and CS speech utterances.
\begin{table}[h]
    \centering
    \caption{ASR performance comparison in terms of substitution, deletion, insertion and MER on the eval set and some examples.}
    \begin{tabular}{l|c|c|c|c}
        \toprule
         \bf{System}&\bf{Sub.}&\bf{Del.}&\bf{Ins.}&\bf{MER}  \\
         \hline
         Baseline&13.6&\bf{4.8}&4.0&22.4\\
         +text from S2S &13.1&5.1&3.9 &22.1\\
         +text from CycleGAN&\bf{13.1}&5.1&\bf{3.7}&\bf{21.9}\\
         \bottomrule
    \end{tabular}
    \label{tab:analysis}
\end{table}
\vspace{0pt}
\begin{minipage}[!htb]{0.5\textwidth}%
\begin{CJK*}{UTF8}{gbsn}
%\begin{table}[!hbt]
    \footnotesize
    %\centering
    %\caption{Some ASR output examples that illustrate the improvement in recognition by adding artificially generated CS text}
    \label{tab:example5}
    \begin{tabular}{l|l}
    \toprule
    \textbf{Reference (EN example)} & SHARED MICROSCOPE\\
    \hline
    \textbf{System} & \\
    \hline
    Baseline & SHARED \underline{\color{red}MY CROSCORD \hspace{1mm}吗}\\
    +text from S2S & SHARED \underline{\color{red}MY CROSCORD \hspace{1mm}吗}\\
    +text from CycleGAN & SHARED MICROSCOPE\\
    \hline
    \hline
    \textbf{Reference (CS example)} & 不\footnote{The pronunciation (IPA) of Mandarin character "不" is \textipa{[bu]}}\hspace{1mm} concern about this\\
    \hline
    \textbf{System} & \\
    \hline
    Baseline & \underline{\color{red}book} concern about this\\
    +text from S2S & \underline{\color{red}book} concern about this\\
    +text from CycleGAN & 不\hspace{1mm} concern about this\\
    \hline
    \hline
     \textbf{Reference (ZH example)} & 不\hspace{1mm} 厉\hspace{1mm} 害\hspace{1mm} 打\hspace{1mm} 架\hspace{1mm} 的\\
    \hline
    \textbf{System} & \\
    \hline
    Baseline & 不\hspace{1mm} \underline{\color{red}利 \hspace{1mm} 太\hspace{1mm} 大\hspace{1mm} 家} 的\\
    +text from S2S & 不\hspace{1mm} \underline{\color{red}利 \hspace{1mm} 太\hspace{1mm} 大\hspace{1mm} 家}\hspace{1mm} 的\\
    +text from CycleGAN & 不\hspace{1mm} 厉\hspace{1mm} 害\hspace{1mm} \underline{\color{red}大\hspace{1mm} 家}\hspace{1mm} 的\\
    \bottomrule
    \end{tabular}
\end{CJK*}
\end{minipage}

\section{Conclusion}
\label{sec:conclusion}
We explored a data augmentation method based on CycleGAN to generate CS text with the goal of improving language modeling and further ASR performance. We considered CS as a speaking style and extended the CycleGAN framework to convert monolingual text to CS text. Our experiments showed that we could generate CS text that not only mimics the CS behaviours in terms of different levels of mixing languages but also the speaking styles in the SEAME corpus. Furthermore, our experiments on SEAME show promising results with consistent improvements in perplexities for language modeling as well as MERs for ASR. Future work could explore in depth differences between the real CS corpus and the artificially generated one with respect to linguistic properties.
%Another interesting effect, namely that the model seems to learn specific speaking styles such as stuttering, will be further investigated in future research.

\clearpage

\bibliographystyle{IEEEtran}
\bibliography{mybib}
\end{document}